\documentclass[10pt,twocolumn,letterpaper]{article}

\usepackage[pagenumbers]{cvpr} % To force page numbers, e.g. for an arXiv version

\usepackage{tikz}
\usepackage{multirow}%

\newcommand{\method}[0]{PanSR} % 

\newcommand{\rotVert}[1]{\parbox[t]{2mm}{\multirow{3}{*}{\rotatebox[origin=c]{90}{#1}}}}

\definecolor{gold}{HTML}{FBF2D2}
\definecolor{silver}{HTML}{DDDDDD}
\definecolor{bronze}{HTML}{EED2B8}

\definecolor{goldD}{HTML}{D9AE13}
\definecolor{silverD}{HTML}{909090}
\definecolor{bronzeD}{HTML}{9A5F26}

\definecolor{catGreen}{HTML}{238763}
\definecolor{catBlue}{HTML}{1F70AE}

\newcommand{\medal}[3]{\tikz[baseline=(char.base)]{\node[rounded corners=2pt,fill=#1,draw=#2,inner sep=1.5pt] (char) {#3};}}

\newcommand{\bm}[2]{
    \ifcase#1\or% case 1
      {\medal{gold}{goldD}{\textbf{#2}}}
    \or % case 2
      {\medal{silver}{silverD}{#2}}
    \or % case 3
      {\medal{bronze}{bronzeD}{#2}}
    \else % default case
      #2
    \fi\ignorespaces
}

\definecolor{cvprblue}{rgb}{0.21,0.49,0.74}
\usepackage[pagebackref,breaklinks,colorlinks,allcolors=cvprblue]{hyperref}

\def\paperID{16321} % *** Enter the Paper ID here
\def\confName{CVPR}
\def\confYear{2025}

\title{PanSR: An Object-Centric Mask Transformer for Panoptic Segmentation}

\author{Lojze Žust, Matej Kristan\\
University of Ljubljana\\
Večna pot 113, 1000 Ljubljana, Slovenia\\
{\tt\small \{lojze.zust, matej.kristan\}@fri.uni-lj.si}
}

\begin{document}
\maketitle
\begin{abstract}

Panoptic segmentation is a fundamental task in computer vision and a crucial component for perception in autonomous vehicles. Recent mask-transformer-based methods achieve impressive performance on standard benchmarks but face significant challenges with small objects, crowded scenes and scenes exhibiting a wide range of object scales. 
%
% ALTERNATIVE:
We identify several fundamental shortcomings of the current approaches: (i) the query proposal generation process is biased towards larger objects, resulting in missed smaller objects, (ii) initially well-localized queries may drift to other objects, resulting in missed detections, (iii) spatially well-separated instances may be merged into a single mask causing inconsistent and false scene interpretations. To address these issues, the we rethink the individual components of the network and its supervision, and propose a novel method for panoptic segmentation PanSR.
%
%We identify several fundamental shortcomings of the current approaches and propose a novel method for monocular panoptic segmentation PanSR, summarized by the following main contributions: (i) an object-centric approach to proposal generation to address the issue of scale imbalance in query selection, (ii) a new proposal-aware matching scheme to prevent query drifting, (iii) a reformulation of mask prediction for thing classes as a localized segmentation task, improving instance separation, and (iv) a new ground-truth matching scheme that synergizes more effectively with the query selection approach.
%
%As a result, 
PanSR effectively mitigates instance merging, enhances small-object detection and increases performance in crowded scenes, delivering a notable +3.4 PQ improvement over state-of-the-art on the challenging LaRS benchmark, while reaching state-of-the-art performance on Cityscapes. 
%By addressing these key challenges, PanSR brings panoptic segmentation closer to practical use in autonomous systems. 
The code and models will be publicly available on \href{https://github.com/lojzezust/PanSR}{GitHub}.

% but largely ignore the inherent differences between background (stuff) segmentation and foreground (thing) instance segmentation tasks that comprise panoptic understanding. In this work we propose a novel method for panoptic segmentation, informed by the observed discrepancies. Our work can be summarized by the following contributions: 1) we disentangle the decoding of background and foreground objects, 2) introduce an instance-based foreground proposal generator, 3) locality constraints for improved instance segmentation and 4) new GT matching algorithm for the multi-scale setting. Our method achieves state-of-the-art performance on panoptic benchmarks for surface navigation and urban navigation and significantly reduces the amount of errors in scenes with large numbers of objects or multiple objects with similar appearance.
\end{abstract}    
\section{Introduction}

% Problem 1: Metode nagnjene k detekciji velikih objektov (zaradi top K)
% Problem 2: 
% Problem 3: loss oz. matching je nenaraven (proposali morajo naslavljati objekte, ki niso v proposalih, tekmujejo med sabo)
% 

\begin{figure}[th]
\centering
\includegraphics[width=\linewidth]{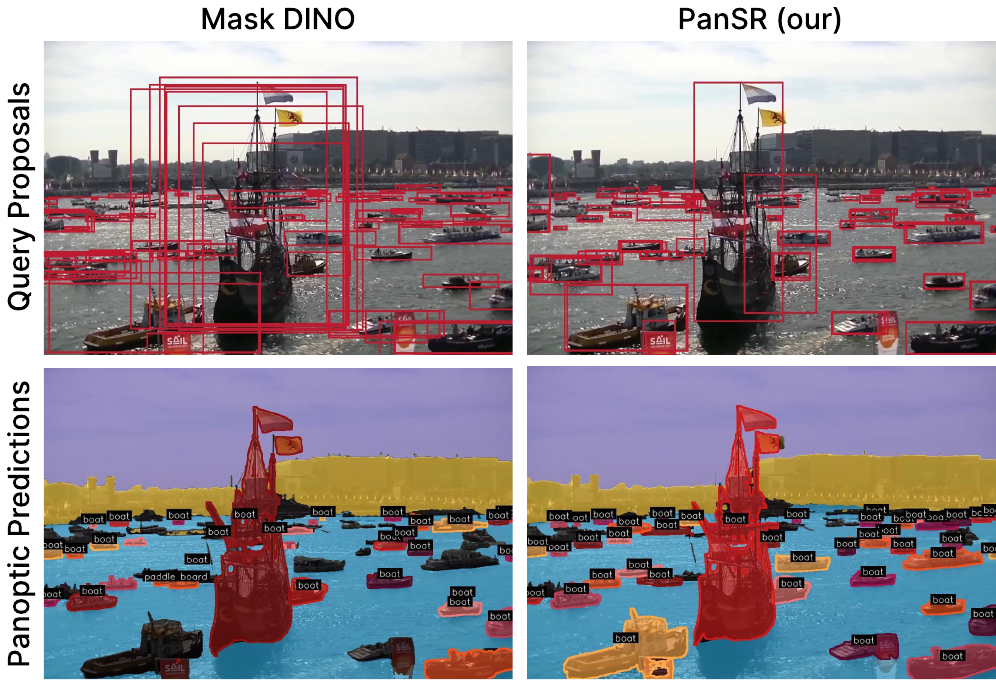}
\caption{Recent transformer-based methods for panoptic segmentation rely on a simple query proposal approach (top left) and struggle with instance separation (bottom left). \method{} presents object-centric query proposals (top right) and reworks the mask decoding process for thing classes, leading to significant improvements in instance segmentation (bottom right).}\label{fig:cover}
% \caption{We introduce a robust panoptic method, that addresses the shortcomings of recent transformer-based methods, including proposal drift during decoding (1) and over-segmentation of similar objects (2).}\label{fig:cover}
\end{figure}

Panoptic segmentation is a fundamental scene understanding problem that jointly addresses semantic segmentation and object detection/instance segmentation tasks. 
Recent advances have led to the emergence of mask transformer architectures~\cite{Cheng2021MaskFormer,Cheng2021Mask2Former,Li2022Mask} with the potential to handle these tasks in a unified architecture.

Mask transformers~\cite{Cheng2021Mask2Former,Li2022Mask,Jain2023OneFormer} perform panoptic segmentation through a set of learnable queries, each encoding an individual mask instance. These queries are gradually refined via cross-attention with multi-scale image features in the transformer decoder. From the final refined queries, object label, mask and optionally a bounding box are predicted. During training, bipartite matching is used for optimal matching between learnable queries and ground-truth instances. This leads to excellent performance on general panoptic benchmarks such as COCO~\cite{Caesar2016COCOStuff} and ADE20k~\cite{Zhou2019Semantic}, as well as scene parsing in autonomous vehicles Cityscapes~\cite{Cordts2016Cityscapes}. However, several shortcomings were exposed on the recent maritime scene understanding benchmark for autonomous boats LaRS~\cite{Zust2023LaRS}. A crucial difference from the related autonomous cars scenes is that the visibility range is substantially larger in maritime scenes. This leads to scenes with a much higher diversity of observed objects' sizes.

% Issue 1: proposals fail to addresss objects of different scales
Recent advances in the detection capabilities of mask transformers, replacing learnable queries with query selection (\ie proposals) mechanisms~\cite{Li2022Mask,Zhang2023Simple}, improve the robustness of models to different object scales. However, we observe that the simple top-k response selection process used in these methods is biased towards large objects due to their relatively larger pixel count, often leading to missed detections of small objects in the same scene (see Figure~\ref{fig:cover}).

\begin{figure}[t]
\centering
\includegraphics[width=\linewidth]{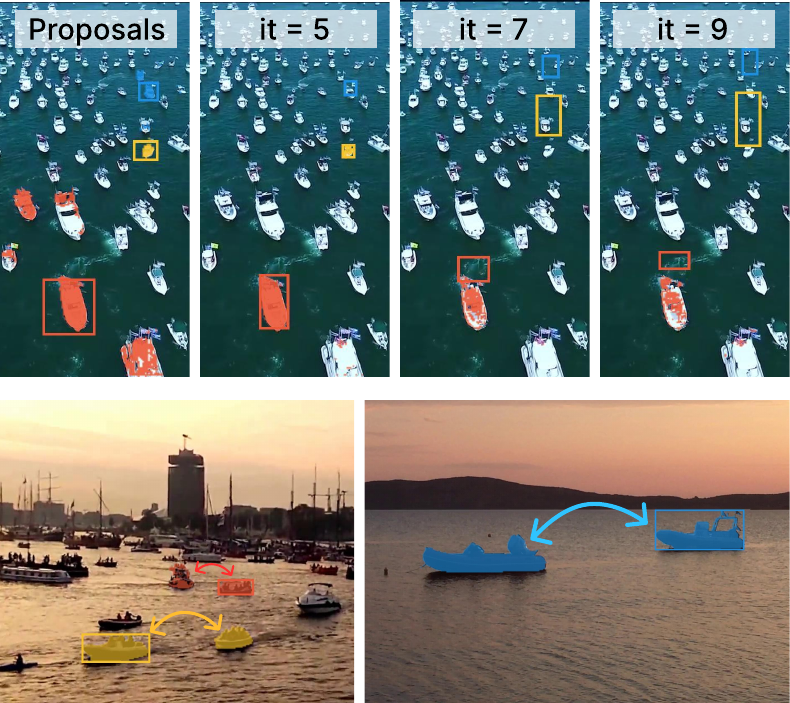}
\caption{Failure cases of mask transformers: well-initialized queries drift away from original objects during decoder iterations (top), and well-separated objects become merged by predicted segmentation masks (bottom). Colors indicate instance labels.}\label{fig:problems}
\end{figure}

% Issue 2: query drift. Potrebno skrajšati
Another issue is \textit{query drift}. The bipartite matching used in training encourages the decoding of matched queries towards the corresponding ground-truth, while pushing the unmatched queries away. However, in case of several proposals generated on the same object, one query will be favored to match the object, while the others tend to be pushed toward other objects. This leads to unpredictable behavior at test-time, where well-generated initial queries drift away from the object during the decoding, even in the absence of alternative queries for the same object (see Figure~\ref{fig:problems}).

% Issue 3: over segmentation (ravnokar izvedel, da je to v resnici undersegmentation, tako da mogoče bolje najti boljši izraz).
% Issue 3: instance merging
Lastly, instance masks are computed by correlating the decoded query with high-resolution image features. This requires learning of strong features to distinguish instance of visually similar objects in feature space. In practice, however, the resulting masks may merge several spatially well-separated instances (Figure~\ref{fig:problems}), even when bounding boxes are correctly localized. This is especially common in crowded scenes.  

% Our method PanSR
To address the aforementioned issues, we propose \method{}, a new scale-robust mask transformer for panoptic segmentation, which is our primary contribution. The method introduces several design novelties: \textbf{(i)} To capture objects at all scales, we introduce an object-centric proposal module (OCP), which shifts proposal extraction from pixel level to object level and improves performance on small objects. \textbf{(ii)} To aleviate query drift, we introduce a new proposal-aware matching scheme, which prevents matching of proposals to incorrect ground-truth instances, and allows multiple proposals to match to a single ground-truth instance. This removes the competition between the proposals of the same object and instead encourages them to decode alternative predictions for the same object. \textbf{(iii)} To address the problem of instance merging, we introduce object-centric mask prediction, which constrains the mask prediction by the predicted bounding boxes. This removes the requirement of learning strong global instance separation features and frees up the network capacity for other capabilities. \textbf{(iv)} Finally, to improve robustness to variance in the proposal extraction process, we introduce a set of mask-conditioned queries during training, by sampling queries from random locations in object regions.

% TODO: 

\begin{figure*}[th]
\centering
\includegraphics[width=0.8\textwidth]{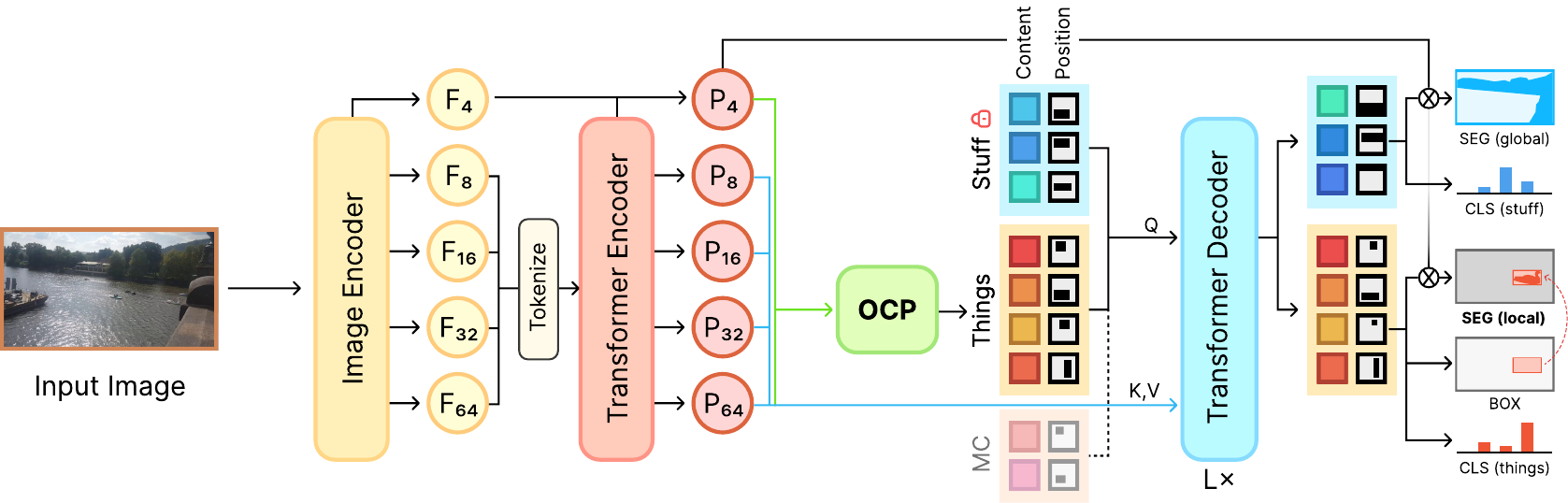}
\caption{Architecture of \method{}. The backbone features are processed by a transformer encoder into a feature pyramid. Object-Centric Proposal extractor (OCP) is used to obtain thing queries. A transformer decoder refines the instance queries. 
The predicted masks of thing classes are limited by their predicted bounding boxes.
%For the thing classes we limit the segmentation prediction to a local area predicted by the bounding box. 
Mask-conditioned queries (MC) ensure robustness to noise in proposal extraction.}\label{fig:architecture}
\end{figure*}

PanSR outperforms all state-of-the-art methods by a large margin (+3.4\% PQ) on the recent challenging LaRS~\cite{Zust2023LaRS} benchmark and performs on par with the best methods on Cityscapes~\cite{Cordts2016Cityscapes} (67.3 PQ). Quantitative and qualitative analysis reveals superior performance, primarily on small objects, in dense scenes, and in scenes with many similar objects.

\section{Related Work}

% Mask transformer architectures (mask2former, max-deeplab)
% Merging of segmentation methods (instance, semantic, panoptic) and detection (DETR, DN-DETR, DINO) -> MaskDINO
% Separation of bg, fg (OpenSeeD), open-vocabulary for bridging the gap between datasets with different labels
% Proposals? Query selection

Panoptic segmentation emerged as a natural merging of semantic segmentation and instance segmentation tasks. This is also reflected in the early methods~\cite{Kirillov2019Panoptic,Kirillov2019Panoptica,Cheng2020Panoptic,Xiong_2019_UPSNet}, which feature a two-branch (semantic and instance) approach with a post-processing fusion of the two outputs.

However, most recent approaches follow a mask-transformer architecture~\cite{Cheng2021MaskFormer,Cheng2021Mask2Former,Wang2020MaXDeepLab}, which reformulates all segmentation tasks (semantic, instance, panoptic) as mask prediction and classification. Inspired by detection transformers~\cite{Carion2020EndtoEnd}, they use a set of instance queries, where each query is decoded into a single instance mask and corresponding class label. 

Recent efforts focus on bridging the gap between panoptic segmentation and detection, by including a bounding box regression head and integrating key insights from recent improvements in detection transformers such as replacing learnable queries with query selection~\cite{Zhu2020Deformable,Zhang2022DINO} and de-noising queries~\cite{Li2022DNDETR,Liu2021DABDETR,Zhang2022DINO} to mask-transformers~\cite{Li2022Mask}. While this leads to much better detection performance, the process selects queries at top $k$ per-pixel score prediction locations, which
%selection process is a simple pixel-level top-k selection mechanism, which 
inherently favors larger objects. Furthermore, with the introduction of these changes, other design choices of mask transformers, which are optimized for learnable queries, have not been reconsidered.

The inherent differences between thing and stuff classes have only been explored recently. A more natural separation into stuff and thing queries were proposed~\cite{Zhang2023Simple}, where learnable queries are used for semantic stuff classes, and query selection is used for thing class instances. We adopt a similar strategy in PanSR.

Several recent approaches integrate vision-language models in an open-vocabulary~\cite{Zhang2023Simple,Xu2023OpenVocabulary} or a multi-task~\cite{Jain2023OneFormer} strategy, removing the constraint of fixed class-sets or tasks. These approaches achieve excellent results by enabling training on multiple datasets or tasks without need for explicit class alignment, but do not address the fundamental issues of mask-transformers tackled in this paper. In this respect, the method we propose is complementary to the vision-language model integration.

\section{\method{}}

At a high level, our method (see Figure~\ref{fig:architecture}) follows a mask-transformer architecture~\cite{Cheng2021Mask2Former,Li2022Mask}. The input image $\mathbf{I}  \in \mathbb{R}^{H \times W \times 3}$ is encoded by image encoder into 
features $\mathbf{F}_s$, where $s \in \{4,8,16,32,64\}$ denotes the stride of the feature maps. The features are further refined by a transformer encoder self-attending the features across scales and building a multi-scale feature pyramid $\mathbf{P}_s$. 
% TODO: Exception on high-resolution masks

A set of $N$ instance queries are used to represent the potential object instances.
They are composed of content ($\mathbf{Q}_f \in \mathbb{R}^{N \times 256}$) and positional (\ie bounding-box, $\mathbf{Q}_\text{box} \in \mathbb{R}^{N \times 4}$) queries.
Due to the unique characteristics of stuff and thing classes, we further split the queries into $N_\text{st}$ stuff and $N_\text{th}$ thing queries and employ slightly different strategies for their decoding and supervision (see Section~\ref{sec:method/query-initialization}).

%We use a set of instance queries to represent each object instance in the image. They are composed of content ($\mathbf{Q}_f \in \mathbb{R}^{N \times 256}$) and positional (\ie bounding-box, $\mathbf{Q}_\text{box} \in \mathbb{R}^{N \times 4}$) queries, where $N$ denotes the total number of queries. Due to the unique characteristics of stuff and thing classes, we further split the queries into $N_\text{st}$ stuff and $N_\text{th}$ thing queries and employ slightly different strategies for their decoding and supervision.

%As a rule, 
Content and positional queries are iteratively refined in several transformer decoder layers $\mathcal{D}_t$, by attending them with the multi-scale image features, \ie
\begin{equation}
    \mathbf{Q}_f^{t+1},\mathbf{Q}_\text{box}^{t+1} = \mathcal{D}_t(\mathbf{Q}_f^t,\mathbf{Q}_\text{box}^t, \mathbf{P}_s),
\end{equation}
where $t \in [1..L]$ denotes the iteration number. Finally, mask, bounding box and class predictions for each query are obtained via the prediction heads
\begin{align}
\mathbf{M}_t = \mathcal{M}(\mathbf{P}_4, \mathbf{Q}_f^t,\mathbf{Q}_\text{box}^t) && \mathbf{y}_t = \mathcal{C}(\mathbf{Q}_f^t) && \mathbf{b}_t = \mathbf{Q}_\text{box}^t,
\end{align}
producing segmentation masks $\mathbf{M}_t$ and class probabilities $\mathbf{y}_t$ respectively. Bounding box predictions $\mathbf{b}_t$ are obtained directly from $\mathbf{Q}_\text{box}^t$.

\subsection{Query initialization}
\label{sec:method/query-initialization}

Because of fundamental difference between the stuff and thing classes, we employ different strategies for the initialization (i.e., $\mathbf{Q}_f^0,\mathbf{Q}_\text{box}^0$) of each type.
Stuff segmentation mainly requires learning  different semantic concepts, which we capture using a set of $N_
\text{st}$ learnable queries~\cite{Cheng2021Mask2Former}. In contrast, the detection of thing objects also requires the distinction between object instances, which has been shown to improve with the use of query selection (\ie proposals) from image features~\cite{Li2022Mask}. To further addresses the scale imbalance of traditional query selection methods, and effectively capture both small and large objects, we introduce an Object-Centric Proposal extraction module (OCP) to extract $N_
\text{th}$ thing queries. 

\begin{figure}[t]
\centering
\includegraphics[width=\linewidth]{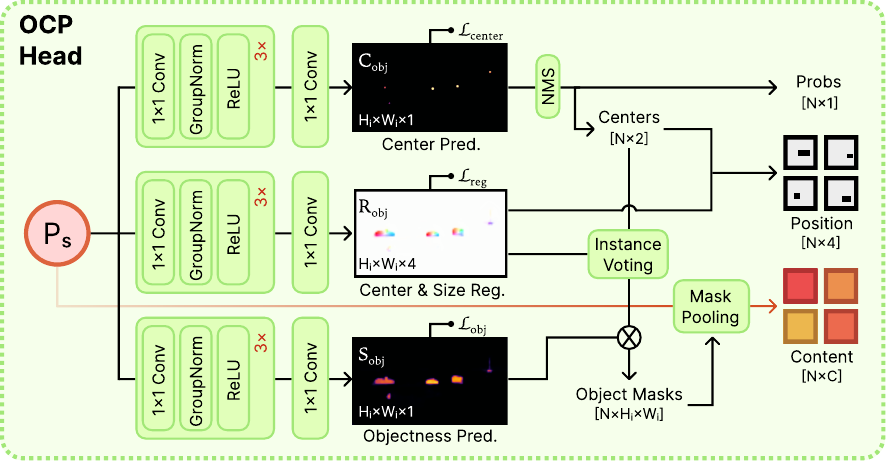}
\caption{Architecture of the OCP module head.}\label{fig:ocp}
\end{figure}

\subsubsection{Object-Centric Proposal Extraction (OCP)}
\label{sec:ocp}

In contrast to traditional pixel-level query selection mechanisms, OCP aims to extract proposals on a per-object level. To achieve this, OCP generates proposals by predicting an object center mask and using non-maxima suppression (NMS) to obtain object centers. On each of the feature levels $\mathbf{P}_s$, an OCP head is built, as summarized in Figure~\ref{fig:ocp}. The OCP head is composed of three small predictions heads. The \textbf{center prediction head}, which serves as the proposal generator, outputs a center mask $\mathbf{C}_\text{obj}$ with high activations at center locations of each object in the scene. %After NMS we
The center locations $x_i$ and their probabilities $p_i$ are obtained by applying a NMS to $\mathbf{C}_\text{obj}$. Proposals are then ranked across all feature levels according to their probabilities $p_i$ and the top $N_\text{th}$ are selected for query extraction.

To extract positional query $\mathbf{Q}_\text{box}^0 (i)$ of object $i$ we utilize a \textbf{regression head}, which predicts object center and size regression maps $\mathbf{R}_\text{obj} \in \mathbb{R}^{H \times W \times 4}$. Specifically, on each object pixel, a relative center location ($\Delta x$, $\Delta y$) and object bounding box size ($w$, $h$) are predicted. The positional queries of can be directly obtained from the regression values at the location $x_i$.

To extract the content query $\mathbf{Q}_\text{f}^0 (i)$, the center location $x_i$ might not be the most reliable. Instead, we rely on an approximate segmentation of the object that can be obtained from the regression maps $\mathbf{R}_\text{obj}$ via instance voting. Each pixel is assigned to the proposal location $x_i$ closest to it's regressed center prediction. The approximate object mask $\mathbf{m}_i$ for the object $i$ at pixel $j$ can be expressed as
\begin{equation}    
m_i(j) =
\begin{cases}
1, & \text{if } d_{i,j} = \min_{k}( d_{k,j} ) \text{ and } d_{i,j} < \theta, \\
0, & \text{otherwise.}
\end{cases}
\end{equation}
where $d_{i,j}$ denotes the distance between a proposal location $x_i$ and the regressed center location (from $\mathbf{R}_\text{obj}$) at pixel $j$. $\theta$ is a distance threshold -- if no proposal locations are within $\theta$ of the predicted center, the pixel is not assigned to any proposal.

%As these rough masks might not be the most reliable, especially towards the edges of the object, 
Since the approximate masks are unreliable, in particular close to object edges,
a third, \textbf{objectness head}
%, is trained, 
%which performs binary segmentation of thing objects
classifies pixels into \textit{thing objects}
(1) vs. \textit{stuff classes} (0), producing a binary segmentation mask $\mathbf{S}_\text{obj}$. To obtain the content query, we then use the probabilities in $\mathbf{S}_\text{obj}$ for weighted mask pooling of image features $\mathbf{P}_s$ inside $m_i$
\begin{equation}
    \mathbf{Q}_f^0 (i) = \sum_j \mathbf{m}_i(j) \, \mathbf{S}_\text{obj}(j) \cdot \mathbf{P}_s (j),
\end{equation}
over all pixel locations $j$.

\paragraph{Supervision.}
Each OCP head is supervised independently via the following combination of losses
\begin{equation}
    \mathcal{L}_\text{OCP}^s = \lambda_\text{obj} \mathcal{L}_\text{obj} + \lambda_\text{reg} \mathcal{L}_\text{reg} + \lambda_\text{center} \mathcal{L}_\text{center}.
\end{equation}
The objectness head is supervised by a \textit{pixel-wise focal loss} $\mathcal{L}_\text{obj}$, the regression head uses \textit{L1 regressions loss} $\mathcal{L}_\text{reg}$ based on ground-truth object bounding boxes, and the center prediction head uses \textit{pixel-wise focal loss} $\mathcal{L}_\text{center}$, where the ground-truth center mask is constructed from small isotropic gaussians at the object center location. 
Additionally, to enforce each feature pyramid level to focus on objects of a specific size range, we only supervise each level by objects within the level's size range. Regions of the masks belonging to other objects are ignored in the loss computation for that level. More details about the OCP supervision are provided in the supplementary.

\subsection{Object-Centric Mask Prediction}
\label{sec:method/local-mask}

% \paragraph{Mask prediction} Mask predictions are formed as a dot product between content queries and a high-resolution feature map ($ \mathcal{M}(\mathbf{q}_i,\mathbf{b}_i) = \mathbf{P}_\frac{1}{4}  \mathbf{q}_i $).
% This essentially computes feature similarity across the entire image and ignores the positional information stored in the positional queries $\mathbf{b}_i$. In order to separate instances of similar-looking objects, the backbone must in turn learn robust feature separation between instances of the same semantic category. This is directly at odds with the classification head $\mathcal{C}(\mathbf{q}_i)$, which has to return the same result for all instances of the same category. To alleviate this, we propose an \textit{object-centric} mask prediction (Section~\ref{sec:method/local-mask}), which constrains the segmentation to the local area of the object of interest. For background queries we employ traditional image-level feature correlation for mask prediction.

As is standard in mask transformer architectures~\cite{Cheng2021Mask2Former}, 
the mask predictions for each \textit{stuff} query is obtained by correlating the respective query feature with the high-resolution image features $\mathbf{P}_4$.
%we employ a dot product between the query features $\mathbf{Q}_f$ and high-resolution mask features $\mathbf{P}_4$ to obtain mask predictions for each \textit{stuff} query. 
However, in the case of instance segmentation on the \textit{thing} classes, such global similarity computation may be detrimental for separation of objects with similar appearance. Instead, we harness the positional information contained in the object positional queries $\mathbf{Q}_\text{box}$. Specifically, we limit the mask predictions of thing classes to the immediate area of the object region prescribed by the predicted bounding box $\mathbf{Q}_\text{box}$. At a pixel position $x$, the mask can be expressed as
\begin{equation}
     \mathbf{M}_t (x) = \begin{cases}
\sigma(\mathbf{P}_4 (x) \cdot  f(\mathbf{Q}_f)) & \text{if } x \text{ in } \phi( \mathbf{Q}_\text{box}, \epsilon_w, \epsilon_h) \\ 
0 & \text{otherwise} \\ 
\end{cases},
\end{equation}
where $\sigma$ is the sigmoid activation function, $f$ is a linear projection layer and and $\phi(\cdot, \epsilon_w, \epsilon_h)$ is a dilation function, that expands the bounding box both directions by a small margins defined by $\epsilon_w$ adn $\epsilon_h$. In other words, the mask probability outside the dilated bounding box is set to zero. The benefits of this approach are two-fold. First, the network avoids learning to separate objects at a global scale and can instead use it's capacity to learn the separation of more challenging, overlapping instances in a local context. Second, this approach explicitly enforces consistency between object segmentation and bounding box prediction, reducing bounding-box-segmentation inconsistencies during matching and inference. 

\subsection{Training}

\begin{figure}[t]
\centering
\includegraphics[width=\linewidth]{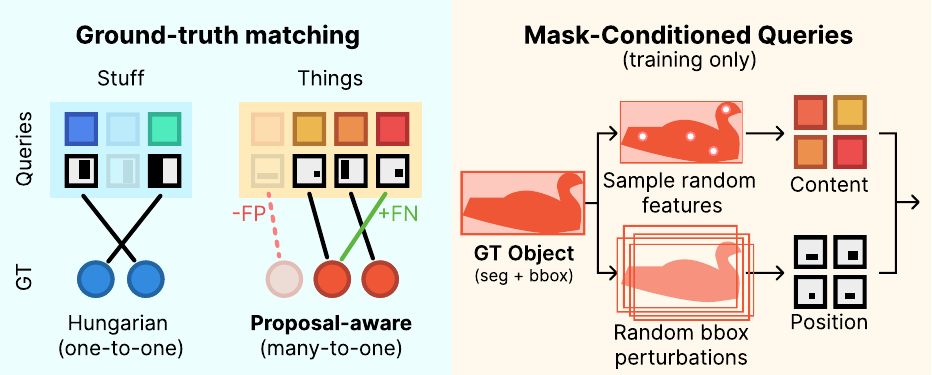}
\caption{The proposal-aware matching scheme (left) alleviates the problem of false-negative and false-positive matches. Mask-conditioned queries (right) simulate random variation in content and positional queries of proposals during training.}\label{fig:training}
\end{figure}

Following the classic training procedure~\cite{Li2022Mask}, we train \method{} by a combination of losses 
\begin{equation}
\mathcal{L}_\text{pred} = \lambda_\text{cls} \mathcal{L}_\text{cls} + \lambda_\text{mask} \mathcal{L}_\text{mask} + \lambda_\text{box} \mathcal{L}_\text{box}
\end{equation}
on matches between queries and ground-truth segments, where $\mathcal{L}_\text{cls}$ is a binary sigmoid focal loss on predicted labels, $\mathcal{L}_\text{mask}$ is a combination of sigmoid cross-entropy and DICE loss on predicted masks, and $\mathcal{L}_\text{box}$ is a combination of L1 and generalized IoU losses on predicted bounding boxes. The total loss is a combination of the proposal and prediction losses
\begin{equation}
    \mathcal{L} = \sum_s \mathcal{L}_\text{OCP}^s + \sum_t \mathcal{L}_\text{pred}^t,
\end{equation}
where OCP is supervised at all feature levels $s$ and predictions are supervised at the output every decoder layer $t$.

For supervising learnable \textit{stuff} queries, we utilize standard Hungarian matching to compute maximal one-to-one matching between ground-truth and predictions. However, we observe such one-to-one matching approach is not ideal for queries constructed from proposals (\ie \textit{things}). For example, the proposal extractor may extract multiple queries for a single objects at different scales. In standard one-to-one matching only one of the queries is matched to the target object, while other queries are punished. Even worse, the unmatched proposals of one object may get matched to a different nearby object instead.

In theory, this should enable the decoder to utilize the redundant queries on one object to address neighboring objects with missing proposals instead. However, we observe in practice that such strategy leads to inconsistent decodings (\ie query drift) from perfectly good query initializations, even when no alternative query proposals for that object exist (see Figure~\ref{fig:problems}). Instead of relying on this inconsistent strategy, we propose an alternative proposal-aware matching scheme, where query decoding is encouraged to stay consistent with initial proposals.

\subsubsection{Proposal-Aware Matching}
\label{sec:method/thing-matching}

% \paragraph{Ground-truth matching} During training, the queries are matched with ground-truth instances for the supervision of mask, bounding box and class predictions. Since we use learnable queries for background segmentation, we use Hungarian matching to obtain the optimal one-to-one matching between the ground truth and queries, as is common practice in such methods~\cite{Cheng2021Mask2Former,Li2022Mask}. For proposal-based foreground queries, we propose a novel proposal-aware matching scheme (Section~\ref{sec:method/thing-matching}), which accounts for the fact that some objects might be missed by the proposals or that multiple proposals may be extracted for a single object.

The proposal-aware matching is posed as a refinement of a standard one-to-one matching. Specifically, we want to remove bad matches (false positives) and add high-quality alternative matches (false negatives) as demonstrated in Figure~\ref{fig:training}. First, maximal bipartite matching between GT and object queries is computed using the Hungarian algorithm, considering the predicted mask, bounding box and class label in the cost as in~\cite{Li2022Mask}. We then apply two refinement stages that address false-positive (FP) and false-negative (FN) matches respectively.

% TODO: Stuff, things: which predictions are used in matching -> move to implementation details?

\paragraph{Stage 1 -- FP removal:} Missing proposals on a ground-truth object can lead to a forced matching to proposals from other objects. This stage aims to remove these bad matches. Specifically, we remove matches with a low overlap of GT and predicted bounding box ($\text{IoU} < \theta_\text{FP} = 0.25$). This prevents learning to move queries from one object to the other during decoding, addressing the query drift.

\paragraph{Stage 2 -- FN correction:} Due to the nature of multi-scale proposal extraction, multiple queries my be extracted for a single object, but bipartite matching will only select one, while others (FNs) will be diminished by the training. This stage aims to identify these alternative queries as %. This is done by identifying 
unmatched queries with a significant overlap with a ground-truth object ($\text{IoU} > \theta_\text{FN} = 0.80$), and add them as additional supervision matches.

Note that this stage largely removes the ability of the decoder to implicitly learn a NMS-like suppression between queries. At test-time we thus apply a classical NMS among detected bounding boxes, considering their predicted confidence scores~\cite{He2017Mask,Pelhan2024DAVE}. 
%To alleviate this, we perform an explicit non-maxima suppression during inference based on predicted bounding boxes and their confidences extracted from label predictions~\cite{NMS v RCNN?}.

% For the background classes proposal-aware matching is not performed and traditional Hungarian matching is used. Only mask and class predictions are considered in the matching cost.

% KONCAL TUKAJ | DONE

\subsubsection{Mask-Conditioned Queries}
\label{sec:method/denoising}

% In theory, the network should be able to decode ... from proposals extracted from any location on the object
% To reduce senistivity to exact proposal locations, we pose mask denoising as ...

%Recall from Section~\ref{sec:?} that the queries are initialized using an approximate object mask estimate 
%Query proposals extract features by masked pooling of object features based on rough masks predicted by the proposal extractor (OCP). However, these initial mask predictions may not be accurate and features may be extracted from different regions of the object as expected. Thus, to make the decoder more robust to noisy proposal feature extraction, we introduce a set of additional queries during training, which simulate varied feature extraction by sampling features at random scales and from random locations inside the ground-truth object masks akin to de-noising approaches~\cite{DN-DINO, MaskDINO}.

The query proposal extraction (OCP from Section~\ref{sec:ocp}) is a noisy process. Thus, to make the decoder more robust to the noise in query extraction, we introduce a set of additional randomized mask-conditioned queries during training akin to de-noising approaches~\cite{Li2022DNDETR,Li2022Mask}. Mask-conditioned queries are simulated by sampling features at random scales and from random locations inside the ground-truth object masks. 

In contrast to existing de-noising strategies, which obtain features by perturbing a learned class embedding, this approach more closely simulates test-time proposal extraction process. To initialize the positional query (bounding box) we use perturbed ground-truth bounding boxes as in~\cite{Li2022DNDETR}. Each of these queries are matched and supervised with their respective ground-truth object. %As in \cite{DN-DINO}, 
The mask-conditioned queries only interact with each other in the decoder and not with regular queries.

% NMS
% Foreground on top of background
% ...

% Experiments
% - Ablations
% - LaRS SotA
% - Other datasets?

\section{Experiments}

%We conduct extensive experiments of panoptic segmentation.% quality of our method. 
%Extensive experimental analysis is conducted.
We compare \method{} with state-of-the-art methods on LaRS (Section~\ref{sec:exp/lars}), a challenging recent panoptic maritime benchmark, and on Cityscapes (Section~\ref{sec:exp/cityscapes}). A detailed ablation study of \method{} is then reported in Section~\ref{sec:exp/abl}.

\begin{table*}[htb]
    \setlength{\tabcolsep}{4pt}
    \centering
    {\footnotesize
    \begin{tabular}{lcllcccccccccccccc}
    \toprule
    &  & \multirow{2}{*}{Method} & \multirow{2}{*}{Backbone} & \multicolumn{4}{c}{PQ (\%)} &  & \multicolumn{4}{c}{RQ (\%)} & & \multicolumn{4}{c}{SQ (\%)} \\
     \cmidrule{5-8} \cmidrule{10-13} \cmidrule{15-18}
      & & &  &    All &    Th &    Th$_a$ &  St  &  & All &    Th &    Th$_a$ &  St & & All &    Th &    Th$_a$ &  St \\
    \midrule
    \multirow{8}{*}{\rotVert{Single-stage}} & &  Panoptic Deeplab~\cite{Cheng2020Panoptic} &  ResNet-50 &       34.7 &       13.4 &       33.0 &       91.4 &    &       40.3 &       19.3 &       46.3 &       96.2 &    &       69.5 &       60.0 &       71.3 &       94.9 \\
    \cmidrule{3-18}
    & &  \multirow{2}{*}{Panoptic FPN~\cite{Kirillov2019Panoptica}} &  ResNet-50 &       40.1 &       21.7 &       35.5 &       89.3 &    &       46.9 &       28.6 &       45.9 &       95.8 &    &       73.5 &       66.1 &       77.3 &       93.1 \\
    & &   & ResNet-101 &       38.7 &       19.7 &       35.5 &       89.4 &    &       45.0 &       26.1 &       46.0 &       95.5 &    &       73.6 &       66.1 &       77.1 &       93.5 \\
    \cmidrule{3-18}
     & &   \multirow{2}{*}{Mask2Former~\cite{Cheng2021Mask2Former}} &  ResNet-50 &       37.6 &       17.0 &       27.9 &       92.4 &    &       43.7 &       23.6 &       37.6 &       97.3 &    &       71.3 &       62.4 &       74.2 &       95.0 \\
     &  &   &     Swin-B &       41.7 &       21.8 &       33.6 & \bm3{94.7} &    &       48.5 &       29.7 &       44.6 & \bm2{98.5} &    & \bm2{78.2} & \bm2{71.5} &       75.3 & \bm3{96.2} \\
        \cmidrule{3-18}
    & &   MaX-DeepLab~\cite{Wang2020MaXDeepLab} &      MaX-S &       31.9 &        9.5 &       19.2 &       91.7 &    &       36.1 &       13.4 &       26.0 &       96.6 &    &       71.3 &       62.5 &       73.7 &       94.8 \\
    \cmidrule{3-18}
    &  &     Mask DINO (1S)~\cite{Li2022Mask} &  ResNet-50 &       41.4 &       22.3 &       30.0 &       92.5 &    &       47.1 &       28.9 &       37.9 &       95.8 &    &       75.2 &       67.2 & \bm2{79.2} & \bm2{96.5} \\
    \cmidrule{3-18}
    &  &  OneFormer~\cite{Jain2023OneFormer} &    Swin-L &       35.2 &       12.8 &       27.8 & \bm2{95.0} &    &       39.2 &       17.1 &       36.6 & \bm3{98.2} &    &       74.3 &       65.8 &       76.0 & \bm1{96.7} \\
    \midrule
    \multirow{2}{*}{\rotVert{Two-stage}} & &    \multirow{2}{*}{Mask DINO (2S)~\cite{Li2022Mask}} &  ResNet-50 &       50.1 &       34.2 &       47.7 &       92.5 &    &       59.0 &       45.0 &       61.4 &       96.1 &    &       74.5 &       66.3 &       77.8 & \bm3{96.2} \\
    & &                 &     Swin-L & \bm3{53.9} & \bm3{38.9} & \bm3{54.6} &       93.8 &    & \bm3{63.0} & \bm3{50.1} & \bm3{69.3} &       97.1 &    &       75.6 &       67.7 & \bm3{78.8} & \bm2{96.5} \\
        \cmidrule{3-18}
    &  &                     \multirow{2}{*}{\textbf{\method{}}} &  ResNet-50 & \bm2{54.2} & \bm2{39.3} & \bm2{56.3} & 94.1 &    & \bm2{63.9} & \bm2{51.1} & \bm2{71.6} & 97.8 &    & \bm1{82.6} & \bm1{77.4} &       78.6 & \bm3{96.2} \\
    &  &                      &     Swin-L & \bm1{57.3} & \bm1{43.0} & \bm1{60.4} & \bm1{95.4} &    & \bm1{66.9} & \bm1{55.0} & \bm1{75.8} & \bm1{98.7} &    & \bm3{75.9} & \bm3{68.2} & \bm1{79.7} & \bm1{96.7} \\
    \bottomrule
    \end{tabular}
    }
    \caption{Panoptic segmentation on the LaRS test set. Top three results for each metric are outlined in gold, silver and bronze.}
    \label{tab:sota}
\end{table*}

\subsection{Implementation details}

The architecture of the transformer encoder and decoder follows \cite{Li2022Mask}. %During training 
We set the number of learnable stuff queries $N_\text{st} = 50$, thing queries $N_\text{th} = 250$ and mask-conditioned queries $N_\text{dn} = 100$. 
%We use the same number of stuff and thing queries during inference. 
We set $\theta = 0.02 \cdot w$, where $w$ is the feature-map width at that level. For bounding-box dilation during mask prediction we use $\epsilon_w = \text{min}(0.1 \cdot w, 2), \epsilon_h = \text{min}(0.1 \cdot h, 2)$ pixels, where $(w,h)$ is the size of the bounding box in pixels. We use $L=9$ layers in the transformer decoder. Unless explicitly stated otherwise, we utilize a ResNet-50 backbone in all our experiments and visualizations.

We adopt the training strategy (losses, LR scheduling, optimizer) from \cite{Li2022Mask}. We set the loss weights $\lambda_\text{obj}=\lambda_\text{reg}=\lambda_\text{center}=5 $ and $\lambda_\text{cls} = 4$, $\lambda_\text{mask} = \lambda_\text{box} = 5$. For experiments on LaRS, we train \method{} on 4 $\times$ A100 GPUs with a total batch size of 8 images for 90k training iterations. We apply the same training schedule to baseline methods.

%During training we use a copy-paste 
The copy-paste
augmentation~\cite{Ghiasi2021Simple} is used in training, pasting additional thing objects from the dataset at random locations in the image. For LaRS, we paste objects inside the water region. We adopt an identical training strategy for Cityscapes, with the exception of using 8 $\times$ A100 GPUs with a total batch size of 16 images. We do not employ copy-paste augmentations during training on Cityscapes.

% TODO: loss, bbox delta param, threshold for matching

% 4xA100 (LaRS), batch size = 8, 8xA100, batch size = 16
% ResNet 50

% 4.1. Setup We conduct extensive experiments on three settings: a closed-set setting on the COCO detection benchmark (Sec. C.1), an open-set setting on zero-shot COCO, LVIS, and ODinW (Sec. 4.2), and a referring detection setting on RefCOCO/+/g (Sec. 4.3). Ablations are then conducted to show the effectiveness of our model design (Sec. 4.4). We also explore a way to transfer a well-trained DINO to the open-set scenario by training a few plug-in modules in Sec. 4.5. The test of our model efficiency is presented in Sec. I. Implementation Details We trained two model variants, Grounding-DINO-T with Swin-T [32], and GroundingDINO-L with Swin-L [32] as an image backbone, respectively. We leveraged BERT-base [8] from Hugging Face [51] as text backbones. As we focus more on the model performance on novel classes, we list zero-shot transfer and referring detection results in the main text. More implementation details are available in the Appendix Sec. A.

\subsection{Panoptic segmentation on LaRS}
\label{sec:exp/lars}

%We compare our method against 
%The proposed \method{} is compared with
%state-of-the-art panoptic segmentation methods on the challenging LaRS benchmark~\cite{Zust2023LaRS},
LaRS~\cite{Zust2023LaRS} is a maritime, panoptic, scene-understanding benchmark for autonomous surface vessels.
It contains over 4000 visually diverse aquatic scenes split into training, validation and a sequestered test sets. Scenes are labeled with per-pixel panoptic annotations for 4 stuff classes and 8 thing (obstacles) classes with over 21.000 thing objects. A wide range of object sizes and crowded scenes are some of the key challenges of LaRS. We follow a standard evaluation protocol, reporting Panoptic Quality (PQ)~\cite{Kirillov2019Panoptic} on the LaRS test set through the evaluation server~\cite{Zust2023LaRS}. PQ combines the measure for instance recognition quality (RQ), and the measure of segmentation quality (SQ). Class-averaged (All) results are reported, as well as scores on thing (Th) and stuff (St) classes separately. In addition, LaRS also measures class-agnostic performance for thing classes ($\text{Th}_a$), where all thing objects are treated as a single class.

\begin{figure}[t]
\centering
\includegraphics[width=\linewidth]{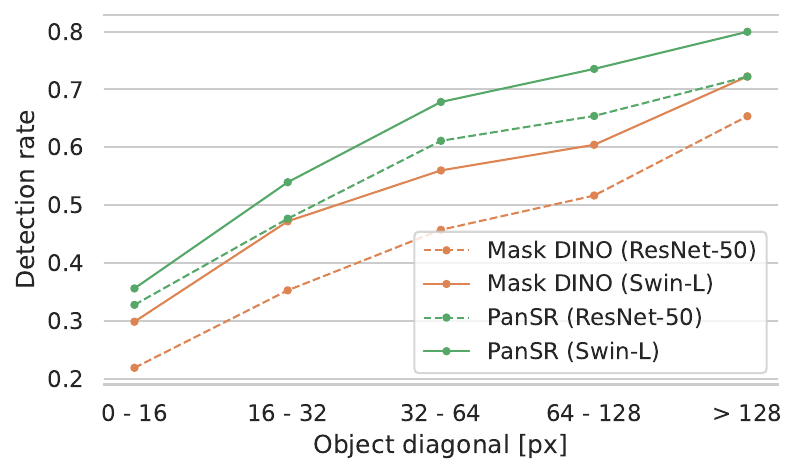}
\vspace{-2em}
\caption{Object detection rate w.r.t. object size.}\label{fig:size_performance}
\vspace{-2em}
\end{figure}

PanSR is compared with the recent state-of-the-art mask transformer methods Mask2Former~\cite{Cheng2021Mask2Former}, MaX-DeepLab~\cite{Wang2020MaXDeepLab}, Mask DINO~\cite{Li2022Mask}, OneFormer~\cite{Jain2023OneFormer}, and with established convolutional methods Panoptic Deeplab~\cite{Cheng2020Panoptic} and Panoptic FPN~\cite{Kirillov2019Panoptica}. 

%The results of the study are presented in
The results are shown in Table~\ref{tab:sota}. Both Mask DINO and \method{} outperform all single-stage methods by a large margin, which demonstrates the importance of query selection for this task. Among the two-stage methods, \method{} achieves a new state-of-the-art by +3.4\% PQ over Mask DINO with Swin-L backbone. Interestingly, \method{} w/ ResNet-50 matches the performance Mask DINO w/ Swin-L, despite a significantly lower backbone complexity, highlighting the effectiveness of its design. 

\begin{figure*}[th]
\centering
\includegraphics[width=\textwidth]{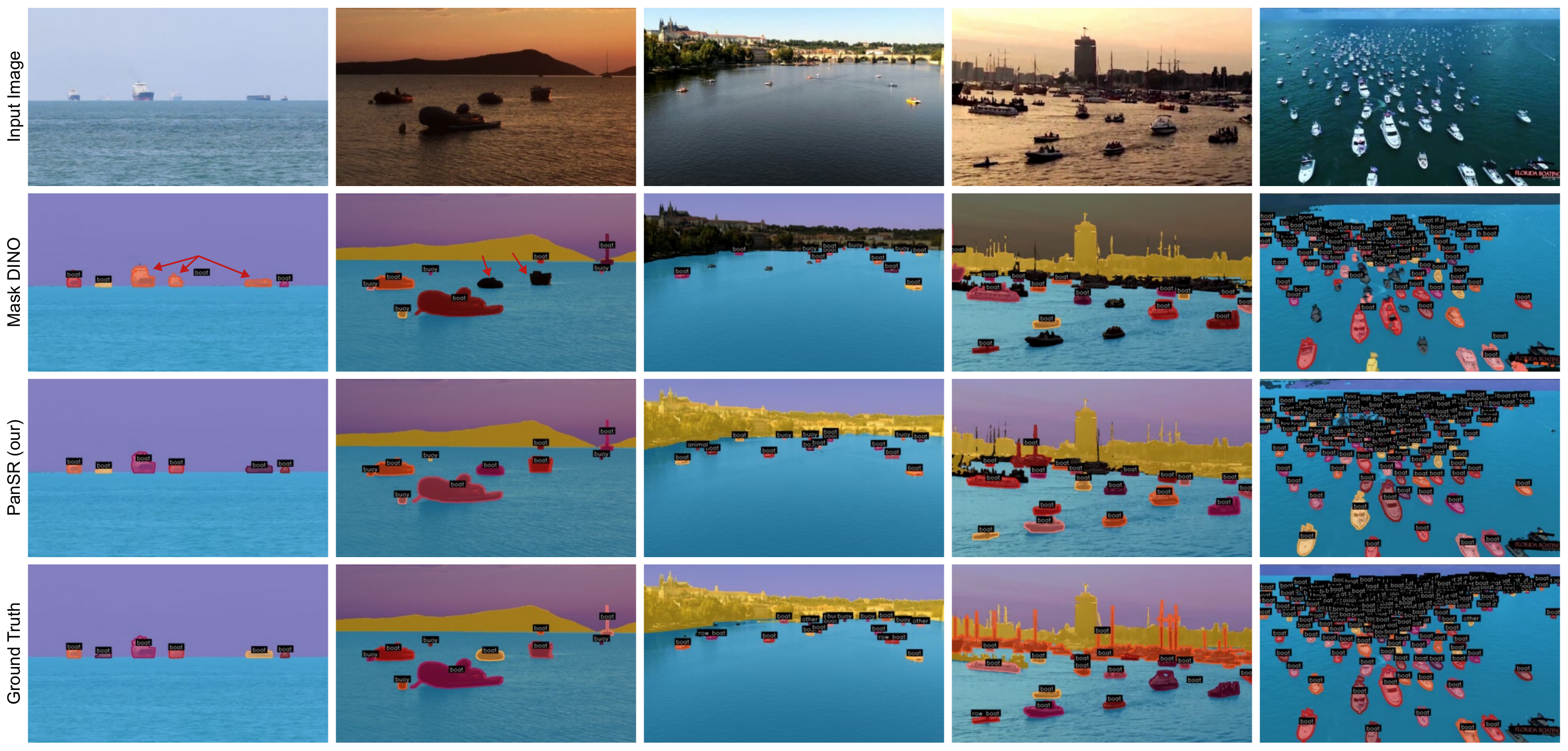}
\vspace{-2em}
\caption{Qualitative results on LaRS. Addressing instance merging (col. 1, 2 \& 4), small objects (col. 3 \& 5) and crowded scenes (col. 5).}\label{fig:lars_qualitative}
\vspace{-1em}
\end{figure*}

Remarkable improvements of \method{}%over other methods can be 
are observed in detection robustness, reflected in the recognition quality metric (RQ). In the thing-classes detection, \method{} outperforms the second-best method by nearly 5\% .
%In detection of thing classes, \method{} improves by almost 5\% over the second-best method. 
This is further confirmed by the qualitative analysis (Figure~\ref{fig:lars_qualitative}), which reveals that \method{} largely eliminates instance grouping of similar objects, facilitating superior detection performance. Compared to other methods, \method{} is also much more accurate in the detection of small objects as shown in Figure~\ref{fig:lars_qualitative}. Additional analysis of detection rate across different object scales, visualized in Figure~\ref{fig:size_performance}, demonstrates that \method{} significantly improves the performance across all instance sizes.

\subsection{Panoptic segmentation on Cityscapes}
\label{sec:exp/cityscapes}

We compare \method{} with the state-of-the-art panoptic segmentation methods on Cityscapes~\cite{Cordts2016Cityscapes}, featuring urban driving scenes. As is common practice, we train \method{} on the training set and report the performance on the val split measured by PQ for panoptic performance, average precision (AP) for detection performance on the thing classes, and mean IoU for semantic segmentation performance.

\method{} is compared with the recent state-of-the-art mask transformer methods Mask2Former~\cite{Cheng2021Mask2Former} and OneFormer~\cite{Jain2023OneFormer}, as well as established convolutional methods Panoptic Deeplab~\cite{Cheng2020Panoptic}, Panoptic FPN~\cite{Kirillov2019Panoptica} and Axial-DeepLab~\cite{Wang2020Axial}. For fair comparison, we include only methods trained exclusively on Cityscapes.
The results are shown in Table~\ref{tab:cityscapes}. 
Using the same parameters as in LaRS and without
dataset-specific hyperparameter optimization,  \method{} reaches a competitive performance among the state-of-the-art methods with a PQ of 67.2, on par with the best method OneFormer~\cite{Jain2023OneFormer}. 

\begin{figure}[th]
\centering
\includegraphics[width=\linewidth]{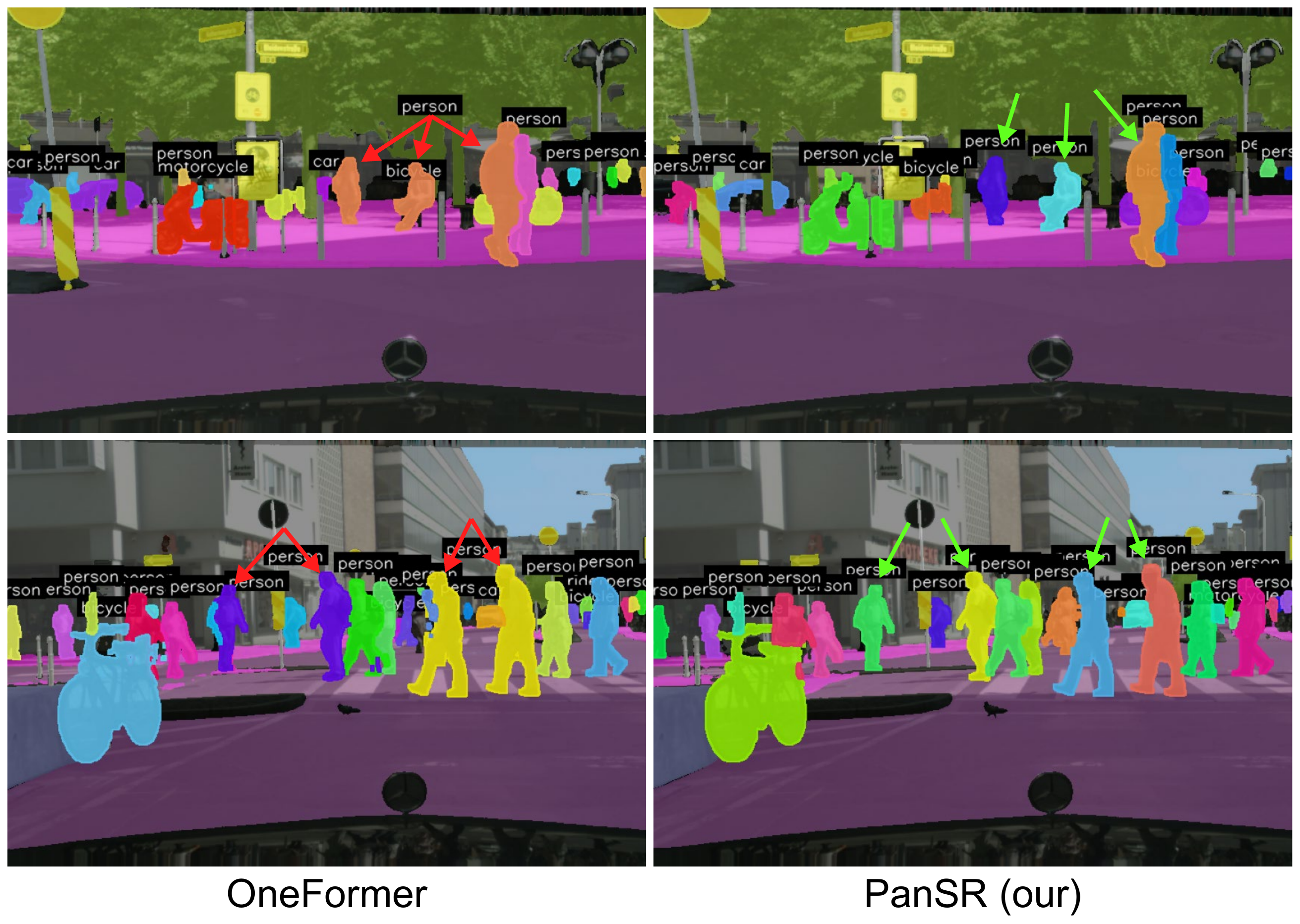}
\caption{Qualitative comparison on Cityscapes val. 
OneFormer (left) incorrectly merges spatially well-separated instances, while PanSR (right) correctly resolves them.
%Examples of instance merging (left) are marked with red arrows.
}\label{fig:cityscapes}
% \caption{We introduce a robust panoptic method, that addresses the shortcomings of recent transformer-based methods, including proposal drift during decoding (1) and over-segmentation of similar objects (2).}\label{fig:cover}
\end{figure}

Note that OneFormer has been trained in a multi-task setup, optimizing for panoptic, instance and semantic segmentation jointly and thus enjoys increased performance on the instance (AP) and segmentation (mIoU) tasks. Qualitative results (Figure~\ref{fig:cityscapes}) confirm that PanSR effectively addresses instance merging issues.
Furthermore, PanSR substantially outperforms OneFormer (by +22.1 PQ) on LaRS, confirming its excellent generalization capabilities.

\begin{table}[]
\setlength{\tabcolsep}{4pt}
\centering
{\small
\begin{tabular}{llccc}
\toprule
Method           & Backbone        & PQ   & AP   & mIoU \\
\midrule
\multirow{2}{*}{Panoptic-DeepLab} & ResNet-50       & 59.7 & -    & -    \\
 & Xception-71     & 63.0 & 35.3 & 80.5 \\
\midrule
Panoptic FPN     & ResNet-50       & 57.7 & 32.0 & 75.0 \\
\midrule
\multirow{2}{*}{Axial-DeepLab}  & Axial ResNet-L  & 63.9 & 35.8 & \bm3{81.0} \\
 & Axial ResNet-XL & \bm3{64.4} & 36.7 & 80.6 \\
\midrule
\multirow{2}{*}{Mask2Former}      & ResNet-50       & 62.1 & 37.3 & 77.5 \\
    & Swin-L          & \bm2{66.6} & \bm2{43.6} & \bm2{82.9} \\
\midrule
OneFormer        & Swin-L          & \bm1{67.2} & \bm1{45.6} & \bm1{84.4} \\
\midrule
% OpenSeed\footnotemark[2]{}      & Swin-L          & \bm1{69.2} & 49.3 & 84.5 \\
\midrule
\multirow{2}{*}{\textbf{\method{}}}      & ResNet-50       & 62.4 &  37.1  & 77.3  \\
    & Swin-L          & \bm1{67.2} & \bm3{43.5}  &  \bm2{82.9}  \\
\bottomrule
\end{tabular}
\footnotetext[\dagger]{Trained on multiple datasets}
}
\caption{Panoptic segmentation results on Cityscapes val.}
\label{tab:cityscapes}
\vspace{-1em}
\end{table}

\subsection{Ablation study}
\label{sec:exp/abl}

\begin{figure*}[th]
\centering
\includegraphics[width=\textwidth]{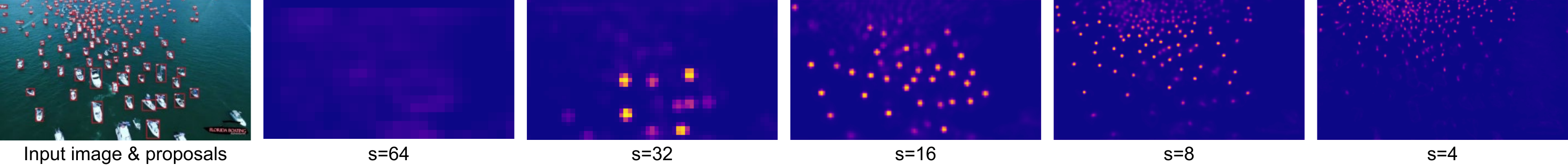}
\vspace{-2em}
\caption{Object center predictions of the OCP module. Each OCP level (denoted by stride $s$) is specialized for objects of specific sizes.}\label{fig:center_pred}
\end{figure*}

We ablate the major contributions of \method{} on the LaRS test set. Results %of the ablation experiments 
are shown in Table~\ref{tab:abl}.  We first investigate the impact of the proposed OCP module (Section~\ref{sec:ocp}) for query selection. Using only learnable queries results in a substantial performance drop on foreground classes. Using the classical query selection mechanisms~\cite{Li2022Mask} dramatically improves performance (+5.7\% PQ). Separation into learnable queries for stuff and proposals for thing classes~\cite{Zhang2023Simple} (denoted \textit{``Lrn. + QS"}) additionally improves performance. Finally, replacing the classical query selection with the proposed OCP module (\textit{``Lrn. + OCP"}) leads to a large PQ improvement of 2.4\%. The difference is especially apparent on the thing classes (+3.4 PQ$_\text{th}$). Figure~\ref{fig:center_pred} illustrates how the OCP head’s center predictions accurately target objects across varying scales.

\begin{table}[t]
    \centering
    {\small
    \begin{tabular}{llccc}
    \toprule
    \textbf{Component} & \textbf{Variant} &    \textbf{PQ} &  \textbf{PQ$_\text{th}$} &  \textbf{aPQ$_\text{th}$} \\
    \midrule
    \multirow{4}{*}{Queries} & Learnable~\cite{Cheng2021Mask2Former} &  43.3 &   24.4 &    36.7 \\
               & Query Selection~\cite{Li2022Mask} &  49.0 &   33.8 &    49.1 \\
               & Lrn. + QS~\cite{Zhang2023Simple} &  51.8 &   36.0 &    52.5 \\
               & \textbf{Lrn. + OCP} &  \textbf{54.2} &   \textbf{39.3} &    \textbf{56.3} \\
    \midrule
    \multirow{3}{*}{Mask} & Global &  52.0 &   36.2 &    51.9 \\
           & Strict BBox &  50.2 &   33.8 &  52.0 \\
               & \textbf{Relaxed BBox} &  \textbf{54.2} &   \textbf{39.3} &    \textbf{56.3} \\
    \midrule
    \multirow{2}{*}{Matching} & Hungarian~\cite{Carion2020EndtoEnd} &  53.3 &   38.2 &    54.4 \\
               % & \textit{Ours} w/o S1 &  52.3 &   36.8 &    55.3 \\
               % & \textit{Ours} w/o S2 &  52.7 &   37.3 &    55.5 \\
               & \textbf{Proposal-aware} &  \textbf{54.2} &   \textbf{39.3} &    \textbf{56.3} \\
    \midrule
    \multirow{3}{*}{\parbox{5em}{Conditioned Queries}} & No &  53.1 &   37.8 &    55.9 \\
               & DN-DETR~\cite{Li2022DNDETR} &  53.9 &   38.9 &    56.2 \\
               & \textbf{Mask-conditioned} &  \textbf{54.2} &   \textbf{39.3} &    \textbf{56.3} \\
    % \midrule
    % \multirow{2}{*}{Copy-paste} & off &  52.7 &   37.5 &    55.1 \\
               % & on &  \textbf{54.2} &   39.3 &    56.3 \\
    \bottomrule
    \end{tabular}
    \caption{Ablations of various components of \method{} on LaRS test set, measured in PQ, PQ$_\text{th}$ on thing classes and class-agnostic aPQ$_\text{th}$.}
    \label{tab:abl}
    }
    \vspace{-1em}
\end{table}

Next, we analyze the importance of object-centric mask prediction (Section~\ref{sec:method/local-mask}). 
%Using traditional global mask predictions works well, but is far from optimal and hinders performance on thing classes. 
Traditional global mask predictions hinder the performance on thing classes.
Applying strict bounding-box constraints to masks of foreground classes is also suboptimal and, in fact, reduces the performance by 1.8\%, as tight bounding boxes interfere with the segmentation of the object. Relaxing these constraints via the proposed bounding box dilation allows additional breathing room for the segmentation, while still constraining it to the immediate area of the object, and results in +2.2\% PQ over global segmentation. The effectiveness of this approach is also evidenced qualitatively by a substantial reduction in instance merging issues as shown in Figure~\ref{fig:lars_qualitative}.

We also investigate our proposal-aware matching scheme (Section~\ref{sec:method/thing-matching}) for thing classes and compare it to traditional one-to-one Hungarian matching. Proposal-aware matching boosts PQ by 0.9\% compared to the traditional approach.
Finally, we consider the impact of mask-conditioned queries during training. In comparison to de-noising approaches~\cite{Li2022DNDETR}, the mask-conditioned queries of \method{} improve performance by 0.3\% PQ.

\section{Conclusion}

We presented PanSR, a novel mask-transformer approach for panoptic segmentation, that addresses key issues of existing approaches such as scale imbalance in query selection, instance merging among visually similar objects, and query drift from good initialization. PanSR achieves this by applying an object-centric approach to proposal generation, mask prediction and ground-truth matching.

%We validate the approach on the challenging LaRS benchmark, featuring a large diversity of different object scale, crowded scenes and small objects. Quantitative evaluation reveals large performance gains across all object size. PanSR also eliminates most cases of instance merging, as demonstrated by our qualitative analysis.

Extensive evaluation of PanSR on the challenging LaRS benchmark, featuring a large diversity of different object scales, crowded scenes and small objects, reveals superior performance across all object sizes and eliminates most cases of instance merging as outlined by our qualitative analysis. PanSR also performs on par with state-of-the-art performance among comparable methods on heavily optimized Cityscapes benchmark, without any additional hyperparameter tuning, demonstrating its good generalization capabilities. By addressing key challenges in scale-robust object detection and instance separation, PanSR brings panoptic segmentation closer to practical use in both autonomous maritime and ground vehicles.

The main contributions of PanSR are complementary to other recent directions in panoptic segmentation such as language-based open-vocabulary~\cite{Zhang2023Simple} mask classification and multi-task learning~\cite{Jain2023OneFormer}. We will explore these avenues in our future work.

\vspace{3em}
\section*{Acknowledgements.}
This work was supported by Slovenian research agency program P2-0214 and project J2-2506, and by a supercomputing network SLING (ARNES, EuroHPC Vega - IZUM).

{
    \small
    \bibliographystyle{ieeenat_fullname}
    \bibliography{main}

\begin{thebibliography}{24}
\providecommand{\natexlab}[1]{#1}
\providecommand{\url}[1]{\texttt{#1}}
\expandafter\ifx\csname urlstyle\endcsname\relax
  \providecommand{\doi}[1]{doi: #1}\else
  \providecommand{\doi}{doi: \begingroup \urlstyle{rm}\Url}\fi

\bibitem[Caesar et~al.(2016)Caesar, Uijlings, and Ferrari]{Caesar2016COCOStuff}
Holger Caesar, Jasper Uijlings, and Vittorio Ferrari.
\newblock {{COCO-Stuff}}: {{Thing}} and {{Stuff Classes}} in {{Context}}.
\newblock \emph{CVPR}, pages 1209--1218, 2016.

\bibitem[Carion et~al.(2020)Carion, Massa, Synnaeve, Usunier, Kirillov, and Zagoruyko]{Carion2020EndtoEnd}
Nicolas Carion, Francisco Massa, Gabriel Synnaeve, Nicolas Usunier, Alexander Kirillov, and Sergey Zagoruyko.
\newblock End-to-{{End Object Detection}} with {{Transformers}}.
\newblock \emph{Lecture Notes in Computer Science (including subseries Lecture Notes in Artificial Intelligence and Lecture Notes in Bioinformatics)}, 12346 LNCS:\penalty0 213--229, 2020.

\bibitem[Cheng et~al.(2020)Cheng, Collins, Zhu, Liu, Huang, Adam, and Chen]{Cheng2020Panoptic}
Bowen Cheng, Maxwell~D Collins, Yukun Zhu, Ting Liu, Thomas~S Huang, Hartwig Adam, and Liang-Chieh Chen.
\newblock Panoptic-{{DeepLab}}: {{A Simple}}, {{Strong}}, and {{Fast Baseline}} for {{Bottom-Up Panoptic Segmentation}}.
\newblock In \emph{Proceedings of the {{IEEE}}/{{CVF Conference}} on {{Computer Vision}} and {{Pattern Recognition}} ({{CVPR}})}, pages 12475--12485, 2020.

\bibitem[Cheng et~al.(2021)Cheng, Schwing, and Kirillov]{Cheng2021MaskFormer}
Bowen Cheng, Alexander~G. Schwing, and Alexander Kirillov.
\newblock Per-{{Pixel Classification}} is {{Not All You Need}} for {{Semantic Segmentation}}.
\newblock \emph{Advances in Neural Information Processing Systems}, 34, 2021.

\bibitem[Cheng et~al.(2022)Cheng, Misra, Schwing, Kirillov, and Girdhar]{Cheng2021Mask2Former}
Bowen Cheng, Ishan Misra, Alexander~G. Schwing, Alexander Kirillov, and Rohit Girdhar.
\newblock Masked-attention {{Mask Transformer}} for {{Universal Image Segmentation}}.
\newblock In \emph{2022 {{IEEE}}/{{CVF Conference}} on {{Computer Vision}} and {{Pattern Recognition}} ({{CVPR}})}, pages 1280--1289, 2022.

\bibitem[Cordts et~al.(2016)Cordts, Omran, Ramos, Rehfeld, Enzweiler, Benenson, Franke, Roth, and Schiele]{Cordts2016Cityscapes}
Marius Cordts, Mohamed Omran, Sebastian Ramos, Timo Rehfeld, Markus Enzweiler, Rodrigo Benenson, Uwe Franke, Stefan Roth, and Bernt Schiele.
\newblock The {{Cityscapes Dataset}} for {{Semantic Urban Scene Understanding}}.
\newblock In \emph{Proceedings of the {{IEEE Conference}} on {{Computer Vision}} and {{Pattern Recognition}} ({{CVPR}})}, 2016.

\bibitem[Ghiasi et~al.(2021)Ghiasi, Cui, Srinivas, Qian, Lin, Cubuk, Le, and Zoph]{Ghiasi2021Simple}
Golnaz Ghiasi, Yin Cui, Aravind Srinivas, Rui Qian, Tsung-Yi Lin, Ekin~D. Cubuk, Quoc~V. Le, and Barret Zoph.
\newblock Simple {{Copy-Paste}} is a {{Strong Data Augmentation Method}} for {{Instance Segmentation}}.
\newblock In \emph{2021 {{IEEE}}/{{CVF Conference}} on {{Computer Vision}} and {{Pattern Recognition}} ({{CVPR}})}, pages 2917--2927, 2021.

\bibitem[He et~al.(2017)He, Gkioxari, Doll{\'a}r, and Girshick]{He2017Mask}
Kaiming He, Georgia Gkioxari, Piotr Doll{\'a}r, and Ross Girshick.
\newblock Mask {{R-CNN}}.
\newblock In \emph{2017 {{IEEE International Conference}} on {{Computer Vision}} ({{ICCV}})}, pages 2980--2988, 2017.

\bibitem[Jain et~al.(2023)Jain, Li, Chiu, Hassani, Orlov, and Shi]{Jain2023OneFormer}
Jitesh Jain, Jiachen Li, MangTik Chiu, Ali Hassani, Nikita Orlov, and Humphrey Shi.
\newblock {{OneFormer}}: {{One Transformer}} to {{Rule Universal Image Segmentation}}.
\newblock In \emph{Proceedings of the {{IEEE}}/{{CVF Conference}} on {{Computer Vision}} and {{Pattern Recognition}}}. arXiv, 2023.

\bibitem[Kirillov et~al.(2019{\natexlab{a}})Kirillov, Girshick, He, and Doll{\'a}r]{Kirillov2019Panoptica}
Alexander Kirillov, Ross Girshick, Kaiming He, and Piotr Doll{\'a}r.
\newblock Panoptic {{Feature Pyramid Networks}}.
\newblock In \emph{Proceedings of the {{IEEE}}/{{CVF Conference}} on {{Computer Vision}} and {{Pattern Recognition}} ({{CVPR}})}, pages 6399--6408, 2019{\natexlab{a}}.

\bibitem[Kirillov et~al.(2019{\natexlab{b}})Kirillov, He, Girshick, Rother, and Dollar]{Kirillov2019Panoptic}
Alexander Kirillov, Kaiming He, Ross Girshick, Carsten Rother, and Piotr Dollar.
\newblock Panoptic segmentation.
\newblock In \emph{Proceedings of the {{IEEE Computer Society Conference}} on {{Computer Vision}} and {{Pattern Recognition}}}, pages 9396--9405. IEEE Computer Society, 2019{\natexlab{b}}.

\bibitem[Li et~al.(2022{\natexlab{a}})Li, Zhang, Liu, Guo, Ni, and Zhang]{Li2022DNDETR}
Feng Li, Hao Zhang, Shilong Liu, Jian Guo, Lionel~M. Ni, and Lei Zhang.
\newblock {{DN-DETR}}: {{Accelerate DETR Training}} by {{Introducing Query DeNoising}}.
\newblock In \emph{Proceedings of the {{IEEE}}/{{CVF Conference}} on {{Computer Vision}} and {{Pattern Recognition}}}, pages 13619--13627, 2022{\natexlab{a}}.

\bibitem[Li et~al.(2022{\natexlab{b}})Li, Zhang, {xu}, Liu, Zhang, Ni, and Shum]{Li2022Mask}
Feng Li, Hao Zhang, Huaizhe {xu}, Shilong Liu, Lei Zhang, Lionel~M. Ni, and Heung-Yeung Shum.
\newblock Mask {{DINO}}: {{Towards A Unified Transformer-based Framework}} for {{Object Detection}} and {{Segmentation}}, 2022{\natexlab{b}}.

\bibitem[Liu et~al.(2021)Liu, Li, Zhang, Yang, Qi, Su, Zhu, and Zhang]{Liu2021DABDETR}
Shilong Liu, Feng Li, Hao Zhang, Xiao Yang, Xianbiao Qi, Hang Su, Jun Zhu, and Lei Zhang.
\newblock {{DAB-DETR}}: {{Dynamic Anchor Boxes}} are {{Better Queries}} for {{DETR}}.
\newblock In \emph{International {{Conference}} on {{Learning Representations}}}, 2021.

\bibitem[Pelhan et~al.(2024)Pelhan, Luke{\v z}i{\v c}, Zavrtanik, and Kristan]{Pelhan2024DAVE}
Jer Pelhan, Alan Luke{\v z}i{\v c}, Vitjan Zavrtanik, and Matej Kristan.
\newblock {{DAVE}} -- {{A Detect-and-Verify Paradigm}} for {{Low-Shot Counting}}.
\newblock \emph{2024 IEEE/CVF Conference on Computer Vision and Pattern Recognition (CVPR)}, pages 23293--23302, 2024.

\bibitem[Wang et~al.(2020{\natexlab{a}})Wang, Zhu, Adam, Yuille, and Chen]{Wang2020MaXDeepLab}
Huiyu Wang, Yukun Zhu, Hartwig Adam, Alan Yuille, and Liang-Chieh Chen.
\newblock {{MaX-DeepLab}}: {{End-to-End Panoptic Segmentation}} with {{Mask Transformers}}.
\newblock In \emph{Proceedings of the {{IEEE}}/{{CVF Conference}} on {{Computer Vision}} and {{Pattern Recognition}} ({{CVPR}})}, pages 5463--5474, 2020{\natexlab{a}}.

\bibitem[Wang et~al.(2020{\natexlab{b}})Wang, Zhu, Green, Adam, Yuille, and Chen]{Wang2020Axial}
Huiyu Wang, Yukun Zhu, Bradley Green, Hartwig Adam, Alan~L. Yuille, and Liang-Chieh Chen.
\newblock Axial-{{DeepLab}}: {{Stand-Alone Axial-Attention}} for {{Panoptic Segmentation}}.
\newblock In \emph{European {{Conference}} on {{Computer Vision}}}, pages 108--126, 2020{\natexlab{b}}.

\bibitem[Xiong et~al.(2019)Xiong, Liao, Zhao, Hu, Bai, Yumer, and Urtasun]{Xiong_2019_UPSNet}
Yuwen Xiong, Renjie Liao, Hengshuang Zhao, Rui Hu, Min Bai, Ersin Yumer, and Raquel Urtasun.
\newblock Upsnet: A unified panoptic segmentation network.
\newblock In \emph{Proceedings of the IEEE/CVF Conference on Computer Vision and Pattern Recognition (CVPR)}, 2019.

\bibitem[Xu et~al.(2023)Xu, Liu, Vahdat, Byeon, Wang, and De~Mello]{Xu2023OpenVocabulary}
Jiarui Xu, Sifei Liu, Arash Vahdat, Wonmin Byeon, Xiaolong Wang, and Shalini De~Mello.
\newblock Open-{{Vocabulary Panoptic Segmentation}} with {{Text-to-Image Diffusion Models}}, 2023.

\bibitem[Zhang et~al.(2022)Zhang, Li, Liu, Zhang, Su, Zhu, Ni, and Shum]{Zhang2022DINO}
Hao Zhang, Feng Li, Shilong Liu, Lei Zhang, Hang Su, Jun Zhu, Lionel~M. Ni, and Heung-Yeung Shum.
\newblock {{DINO}}: {{DETR}} with {{Improved DeNoising Anchor Boxes}} for {{End-to-End Object Detection}}, 2022.

\bibitem[Zhang et~al.(2023)Zhang, Li, Zou, Liu, Li, Gao, Yang, and Zhang]{Zhang2023Simple}
Hao Zhang, Feng Li, Xueyan Zou, Shilong Liu, Chunyuan Li, Jianfeng Gao, Jianwei Yang, and Lei Zhang.
\newblock A {{Simple Framework}} for {{Open-Vocabulary Segmentation}} and {{Detection}}, 2023.

\bibitem[Zhou et~al.(2019)Zhou, Zhao, Puig, Xiao, Fidler, Barriuso, and Torralba]{Zhou2019Semantic}
Bolei Zhou, Hang Zhao, Xavier Puig, Tete Xiao, Sanja Fidler, Adela Barriuso, and Antonio Torralba.
\newblock Semantic {{Understanding}} of {{Scenes Through}} the {{ADE20K Dataset}}.
\newblock \emph{Int J Comput Vis}, 127\penalty0 (3):\penalty0 302--321, 2019.

\bibitem[Zhu et~al.(2020)Zhu, Su, Lu, Li, Wang, Dai, and Research]{Zhu2020Deformable}
Xizhou Zhu, Weijie Su, Lewei Lu, Bin Li, Xiaogang Wang, Jifeng Dai, and Sensetime Research.
\newblock Deformable {{DETR}}: {{Deformable Transformers}} for {{End-to-End Object Detection}}.
\newblock 2020.

\bibitem[{\v Z}ust et~al.(2023){\v Z}ust, Per{\v s}, and Kristan]{Zust2023LaRS}
Lojze {\v Z}ust, Janez Per{\v s}, and Matej Kristan.
\newblock {{LaRS}}: {{A Diverse Panoptic Maritime Obstacle Detection Dataset}} and {{Benchmark}}.
\newblock In \emph{Proceedings of the {{IEEE}}/{{CVF International Conference}} on {{Computer Vision}}}, pages 20304--20314, 2023.

\end{thebibliography}
}

% WARNING: do not forget to delete the supplementary pages from your submission 
\clearpage
\setcounter{page}{1}
\maketitlesupplementary

\begin{figure*}[th]
\centering
\includegraphics[width=\textwidth]{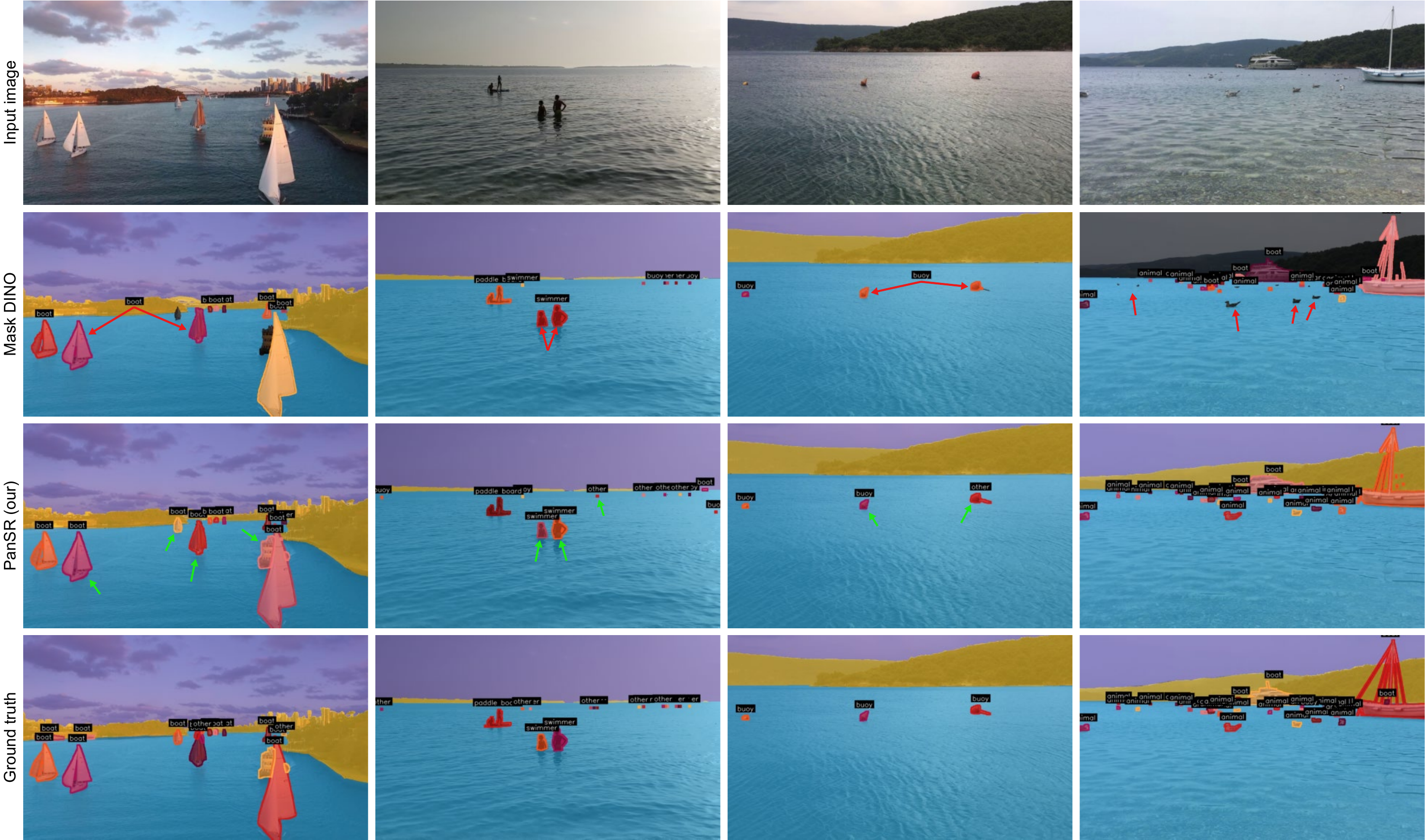}
\caption{Additional qualitative examples on the LaRS test set.}\label{fig:sup/lars_qualitative}
\end{figure*}

\section{OCP supervision details}
\label{sec:sup/ocp}

\subsection{Center prediction ground truth}

To construct the ground-truth center masks $\mathbf{\hat{C}}_\text{obj}$ for supervision of the \textit{center prediciton head} (see Section~\ref{sec:ocp}), we utilize the following procedure. First, an object center $\mathbf{c}_i = (c_x, c_y)$ is extracted from each object's bounding box. A small isotropic 2D Gaussian is then constructed at the object center location. Specifically, the center mask $\mathbf{\hat{C}}_\text{obj}^i$ of object $i$ is computed as
\begin{equation}
    \mathbf{\hat{C}}_\text{obj}^i (\mathbf{x}) = \text{exp} \left ( -\frac{1}{2} \frac{\| \mathbf{x} - \mathbf{c_i} \|^2}{\sigma^2} \right ),
\end{equation}
where $\mathbf{x} = (x_x, x_y)$ is a pixel location in the mask and $\sigma^2 = 1 \text{px}$ is the variance of the Gaussian function. The same fixed variance is used to construct object masks at all OCP levels. Gaussians for individual objects are combined in the final ground-truth center mask using a maximum operation
\begin{equation}
    \mathbf{\hat{C}}_\text{obj} (\mathbf{x}) = \text{max}_i \left ( \mathbf{\hat{C}}_\text{obj}^i (\mathbf{x}) \right ).
\end{equation}
Examples of center predictions obtained using such supervision are shown in Figure~\ref{fig:center_pred} of the main paper.

\subsection{Targeting specific object sizes}

To ensure balance between object scales, each OCP head (Section~\ref{sec:ocp}) specializes in objects of specific scale. We measure the object scale as the bounding box diagonal $d$ (in pixels of the original image). The scale ranges of each OCP level are presented in Table~\ref{tab:sup/scales}. There is a slight overlap between objects targeted by neighboring levels, to ensure consistent operation for objects whose size is close to the limit of each level.

We wish to extract proposals only for objects within the sizer range of the current level, thus \textit{center mask prediction} at all other object center locations are supervised as 0. On the other hand, for the predictions of the \textit{regression} and the \textit{objectness} heads, we relax the supervision slighlty. Instead of forcing zero-outputs on objects not in the scale range, we ignore their contributions to the loss instead. This way, the heads may learn the simplest strategy for predicting regression and objectness outputs at that scale, without the need to differentiate between objects of different scales.

\begin{table}[t]
    \centering
    \begin{tabular}{cc}
    \toprule
    OCP level & size range ($d$) \\
    \midrule
    $s=64$      & $[256, \infty]$  \\
    $s=32 $     & $[128, 512] $      \\
    $s=16$      & $[64, 256]$        \\
    $s=8$       & $[32, 128]$        \\
    $s=4$       & $[0, 64]$         \\
    \bottomrule
    \end{tabular}
    \caption{Size ranges of objects, that are used for supervision at different OCP levels.}
    \label{tab:sup/scales}
    \vspace{-1em}
\end{table}
\section{Additional qualitative examples}

\begin{figure*}[th]
\centering
\includegraphics[width=\textwidth]{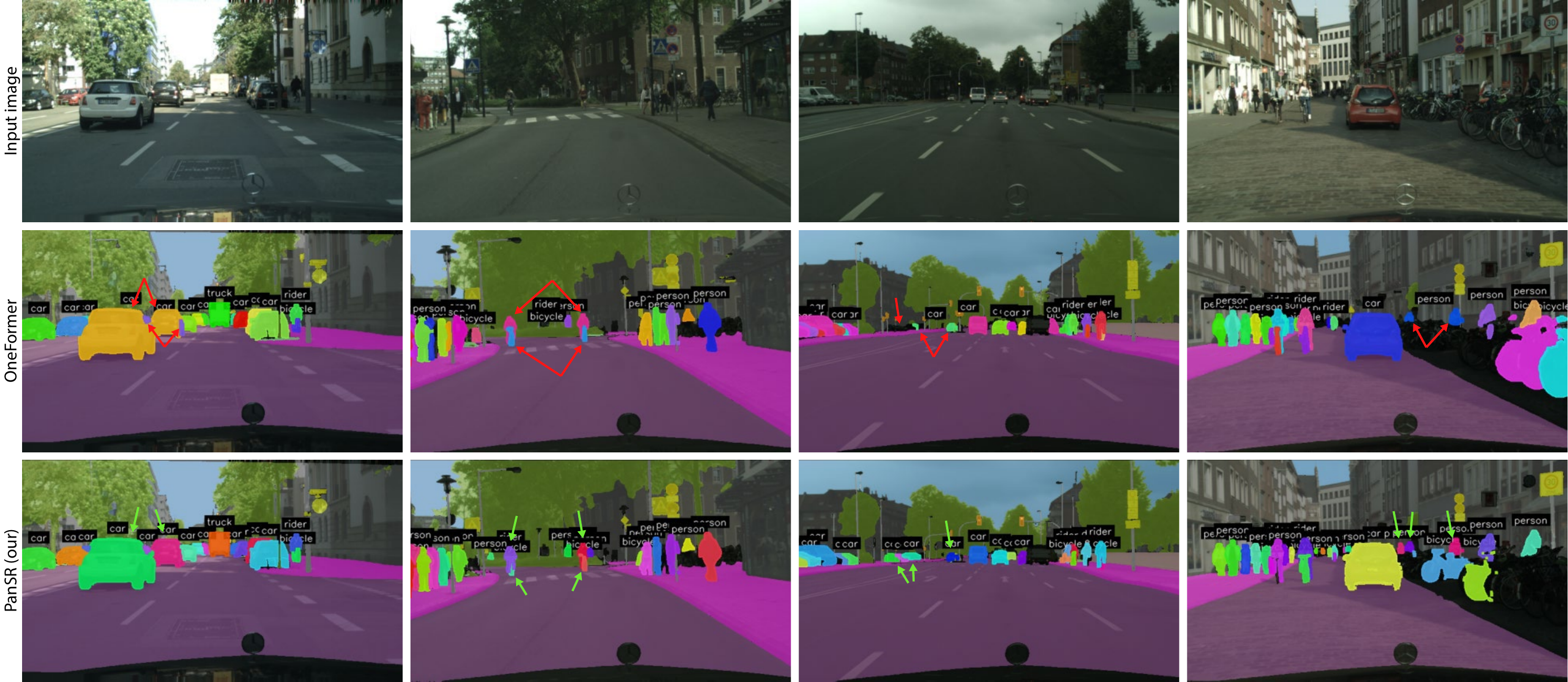}
\caption{Additional qualitative examples on the Cityscapes val set.}\label{fig:sup/cityscapes_qualitative}
\end{figure*}

In Figures~\ref{fig:sup/lars_qualitative} and \ref{fig:sup/cityscapes_qualitative} we present additional qualitative results of PanSR and state-of-the-art methods Mask DINO~\cite{Li2022Mask} and OneFormer\cite{Jain2023OneFormer} on the LaRS test set and Cityscapes val set respectively.

% \section{Video material}

% In the video attached to the supplementary, we further demonstrate the issues of instance merging and query drift in the decoding process (exposed in Figure~2 of the main paper) on practical examples. We also show how PanSR effectively addresses these problems.

\end{document}